\documentclass[twocolumn]{htl-author}

\DOI{2012.0357}

\begin{document}

\title{Seamless Augmented Reality Integration in Arthroscopy: A Pipeline for Articular Reconstruction and Guidance}
\author{Anonymous}
\author{Hongchao Shu$^{1}$, Mingxu Liu$^{1}$, Lalithkumar Seenivasan$^{1}$, Suxi Gu$^{2}$, Ping-Cheng Ku$^{1}$, Jonathan Knopf$^{3}$, Russell Taylor$^{1}$, and Mathias Unberath$^{1}$}
\address{$^{1}$Department of Computer Science, Johns Hopkins University, Baltimore, Maryland, USA\\
$^{2}$Department of Orthopedics, Tsinghua Changgung Hospital,  Tsinghua University, School of Medicine, Beijing, China\\
$^{3}$Arthrex, Inc. Naples, Florida, USA\\
E-mail: hshu4@jhu.edu, unberath@jhu.edu\\}

\historydate{Published in Healthcare Technology Letters}

\abstract{Arthroscopy is a minimally invasive surgical procedure used to diagnose and treat joint problems. The clinical workflow of arthroscopy typically involves inserting an arthroscope into the joint through a small incision, during which surgeons navigate and operate largely by relying on their visual assessment through the arthroscope. However, the arthroscope's restricted field of view and lack of depth perception pose challenges in navigating complex articular structures and achieving surgical precision during procedures. Aiming at enhancing intraoperative awareness, we present a robust pipeline that incorporates simultaneous localization and mapping, depth estimation, and 3D Gaussian splatting to realistically reconstruct intra-articular structures solely based on monocular arthroscope video. Extending 3D reconstruction to Augmented Reality (AR) applications, our solution offers AR assistance for articular notch measurement and annotation anchoring in a human-in-the-loop manner. Compared to traditional Structure-from-Motion and Neural Radiance Field-based methods, our pipeline achieves dense 3D reconstruction and competitive rendering fidelity with explicit 3D representation in \rev{7} minutes \rev{on average}. When evaluated on four phantom datasets, our method achieves $\text{RMSE} = 2.21 \text{mm}$ reconstruction error, $ \text{PSNR} = 32.86$ and $\text{SSIM} = 0.89$ on average. Because our pipeline enables AR reconstruction and guidance directly from monocular arthroscopy without any additional data and/or hardware, our solution may hold \rev{the} potential for enhancing intraoperative awareness and facilitating surgical precision in arthroscopy. Our AR measurement tool achieves accuracy within $1.59 \pm 1.81\text{mm}$ and the AR annotation tool achieves a $\text{mIoU}$ of $0.721$.}

\maketitle

\section{Introduction}

Arthroscopy is a minimally invasive intervention that enables surgeons to examine a joint and repair cartilage, smoothen bone surfaces, or repair ligaments, with reduced surgical trauma and better patient recovery time compared to traditional open surgery. Despite being one of the most common surgical procedures worldwide, knee arthroscopy carries about a $1\%$ risk of complications and accounted for $5\%$ of all pyogenic knee arthritis cases up until 2018~\cite{arthroscopy}. Some complications are associated with tissue or nerve damage resulting from surgeon errors. During the procedure,  visualization can be hindered by poor lighting, obstructions, and inconsistent image quality, posing challenges in obtaining a clear and comprehensive view~\cite{burman1934arthroscopy}. Furthermore, the small incisions used in arthroscopy \rev{limit} access to the joint, making it challenging for surgeons to accurately navigate the surgical site. 
Accurate tissue identification, precise instrument manipulation, and robust navigation~\cite{chen2021augmented} could significantly improve \rev{the} overall success of the procedure and patient recovery. 3D reconstruction and augmented reality (AR) technologies hold the potential to address these issues by complementing the visual input with 3D information, providing better depth cues, and offering real-time guidance, which could improve surgical precision, reduce errors, and facilitate more effective procedures without adding complexity. However, current imaging technologies face limitations: MRI and CT scans are limited for real-time, intraoperative guidance in arthroscopy~\cite{fu2021future}, and traditional RGB scene reconstruction struggles with confined view spaces and limited image features~\cite{8593501}.
To this end, we introduce a pipeline that sequentially resolves 3D arthroscopy scene reconstruction and employs AR tools to enhance surgeons' spatial perception.

In this work, we propose a reconstruction and AR guidance pipeline for arthroscopy that overcomes limitations present in prior systems by achieving high-fidelity 3D reconstruction from limited arthroscope views in confined spaces and enhancing spatial awareness through AR applications. The pipeline incorporates OneSLAM~\cite{Oneslam} to obtain sparse 3D priors and uses a pseudo depth-aided 3D GS model for 3D densification and surface alignment. Leveraging our real-time rendering and high-fidelity reconstruction capabilities, we create AR applications for annotating and measuring critical articular structures. To the best of our knowledge, our approach is one of the first to provide reconstruction and AR tools for arthroscopy assistance based solely on vision input. 
Quantitative and qualitative experiments demonstrate our pipeline's superiority in terms of reconstruction and annotation accuracy compared to both traditional and learning-based
methods.

\section{\rev{Related Works}}
Surgical navigation systems~\cite{chen2021augmented,jeung_augmented_2023,penza2023augmented,shu2023twin,chen2017development,gu2022calibration} have been introduced to aid surgeons with spatial perception. Most of these navigation systems require prior geometric knowledge of the surgical scene and precise real-time tracking of surgical instruments. 
Alternatively, \rev{vision}-based navigation systems empowered by Simultaneous Localization and Mapping (SLAM) algorithms~\cite{marmol_dense-arthroslam_2019,oliva2023orb,Qiu_2018_CVPR_Workshops,gu2023nail} could also be employed for surgical navigation. 
SLAM algorithms enable a vision-based system to navigate in an unknown environment by continuously building a map of the environment while simultaneously keeping track of its location within that map. SLAM algorithms estimate 3D structures and 6 DoF camera poses from 2D image sequences in near video frame rate relying on image feature correspondences across frames~\cite{taketomi2017visual}. Many state-of-the-art SLAM techniques demonstrate spatial-temporal consistency and robustness in \rev{the} general computer vision domain by utilizing either sparse image features~\cite{orb} or direct photometric information~\cite{lsd, dso}. However, the internal structures of the knee often pose unique challenges~\cite{marmol_dense-arthroslam_2019}: they often have similar colors and textures, lacking distinct visual features that feature extractors rely on to distinguish different areas. Additionally, illumination conditions can change drastically with camera movement, further complicating the extraction of consistent photometric information.

In addition to geometric priors generated from the SLAM techniques, a dense 3D model is critical for real-time navigation. It provides surgeons with a continuous and precise map of the joint and facilitates the integration of AR applications. Dense reconstruction techniques, such as Dense-ArthroSLAM~\cite{marmol_dense-arthroslam_2019} leverages Multi-View Stereo (MVS)~\cite{MVS} to reconstruct the knee joint but requires external tracking for 3D prior estimation. 
Neural Radiance Fields (NeRF)-based methods~\cite{nerf} are known for their high-fidelity rendering of non-rigid endoscopic scenes. However, the implicit nature of certain approaches~\cite{wang2022neural,endosurf} and their computational demands present challenges for integration into augmented reality (AR) applications. On the other hand, methods that provide explicit surface representations~\cite{wang2021neus, li2023neuralangelo} require significant time to converge. 3D Gaussian Splatting (3D GS)~\cite{3dgs} based method EndoGS~\cite{endogs} proposed to reconstruct dynamic endoscopic scenes, and achieves superior rendering quality. However, 3D GS models can get trapped in incorrect geometry due to the multi-solution nature of 3D GS~\cite{GRM}, causing floating artifacts which \rev{reduce} the fidelity of the reconstruction~\cite{go}.
Dense reconstruction provides the foundational spatial understanding required for AR applications to deliver realistic, accurate, and interactive augmented experiences. By leveraging detailed 3D models of the environment, the AR system provides a variety of practical scenarios for surgical navigation tasks with a direct interface.~\cite{Kleinbeck2024Jul}
Previous AR guidance systems (Jeung et al.~\cite{jeung_augmented_2023}, Ma et al.~\cite{ma_knee_2020}, Penza et al.~\cite{penza2023augmented}) highlight the importance of accurate overlay and localization, similar to our approach. However, they either rely on bulky tracking devices, produce artificial-looking images, or lack essential AR tools to streamline clinical procedures.

\section{Method}
\begin{figure*}[!t]
\centerline{\includegraphics[width=\textwidth]{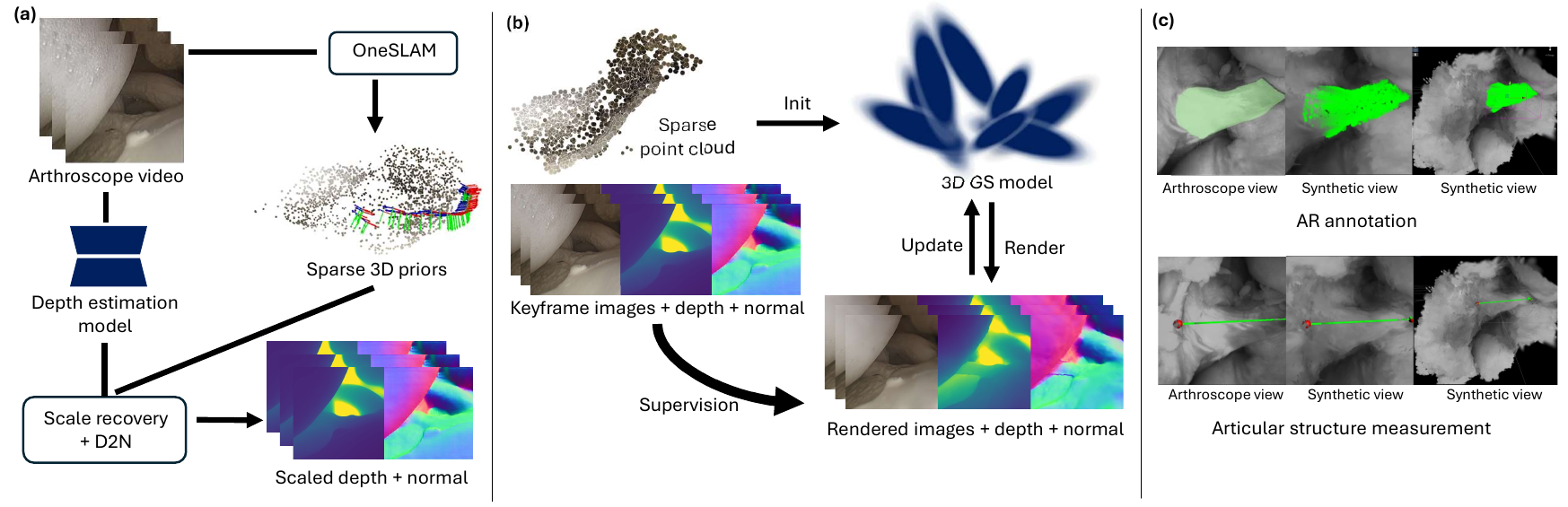}}
\caption{Illustration of the proposed pipeline for reconstruction and AR guidance in Arthroscopy. (a) We reconstruct sparse 3D priors with OneSLAM on an arthroscope footage. For each keyframe, the depth estimation model generates pseudo depth map, the relative scales are recovered for each frame for consistency, normals are generated from a depth-to-normal translator. (b) The sparse point cloud is used to initialize the location of 3D Gaussians, and RGBD images along with normals are taken as supervision for 3D GS model training. (c) In our AR application, we have designed an annotation tool to highlight anatomical structures and a measurement tool to conveniently estimate surface distances.}
\label{fig:pipeline}
\end{figure*}

We introduce a pipeline that incorporates SLAM, monocular depth estimation, and surface reconstruction for dense reconstruction of surgical scenes during arthroscopy. Firstly, we employ OneSLAM~\cite{Oneslam} to reconstruct a sparse 3D point map across multiple keyframes in an arthroscopic video (Sec.~\ref{sec3.1}). Secondly, leveraging a monocular depth estimation model~\cite{depthanything}, we generate frame-by-frame disparity maps, without scale, providing pseudo-depth information. 
We integrate this with sparse 3D correspondences from OneSLAM~\cite{Oneslam} to recover consistent relative depth and generate normal from depth (Sec.~\ref{sec3.2.3}), as shown in \rev{Fig.~\ref{fig:pipeline}(a)}.
Finally, we densely reconstruct a photo-realistic 3D scene leveraging a 3D GS model (\rev{Fig.~\ref{fig:pipeline}(b)}). We enhance 3D GS with geometric supervision and opacity management to better align 3D Gaussians to real articular surfaces (Sec.~\ref{sec3.2.2}). The generated 3D scene facilitates AR applications (Sec.~\ref{sec3.3}), such as AR superimposing and user interaction for articular notch measurement and annotation anchoring (\rev{Fig.~\ref{fig:pipeline}(c)}).


\subsection{OneSLAM Sparse-view Reconstruction}\label{sec3.1}
We exploit the robust point-tracking capabilities of the CoTracker~\cite{cotracker} in OneSLAM~\cite{Oneslam} to overcome the scarcity of distinct visual features on arthroscopic scenes. Image points~\imgpoint{M} are initialized, and each point $x_{i} \in \mathbb{R}^{2}$ is tracked throughout the video~\images{N},\image $\in \mathbb{R}^{H \times W \times 3}$ until it becomes invisible, $N$ and $M$ represent the total number of frames and total number of points, respectively. 
Given a set of matched point correspondences, OneSLAM~\cite{Oneslam} then estimates the relative camera poses~\poseSet{N},~\pose{i}$\in \mathbb{SE}(3)$ using a RANSAC-based PnP algorithm, by minimizing the reprojection error. Additionally, it performs \rev{bundle} adjustments based on keyframes, to reduce the discrepancy between the observed image points and the projected 3D points. Here, the keyframes are selected based on the point set similarity to avoid selecting images with similar view angles. To maintain low computational cost, it adopts a sliding window strategy, jointly optimizing only a subset of sparse point map~\pointMap{K} and camera poses~\poseSet{K} in each window.

\subsection{Pseudo Depth and Normal Generation}\label{sec3.2.3} Depth Anything~\cite{depthanything}, a monocular depth estimation model, is employed to generate initial disparity maps~\disps for keyframes~\keyframes $\subset$ \images{}. The initial depth values are normalized to $0 \sim 1$, to mitigate the effects of random scales. 
As the disparity values inferred for image correspondences fluctuate throughout the video and \rev{restrict} the 3D-GS model from establishing geometric consistency, the mapping between disparities and scaled pseudo-depth is \rev{modeled} as:

\begin{align*}
   D^{pseudo} = \frac{A}{d} + B 
\end{align*}
where $A$ represents the scale that needs to be recovered and $B$ represents a constant shift. 

Based on the 3D priors from OneSLAM~\cite{Oneslam}, we establish 2D-3D correspondences among the disparity ($x_d$), RGB image ($x$), and point map ($X$). We employ the Nelder-Mead algorithm~\cite{nelder1965simplex} to find a function that optimizes the following minimum:
\begin{align*}
    A, B = \underset{A,B}{\arg\min} \left \|X_z - (\frac{A}{x_d} + B)\right \|_{2}
\end{align*}
For all keyframes, unique pairs of $(A, B)$ are obtained to generate pseudo-depth maps with temporal consistency. We then use a depth-to-normal translator D2NT~\cite{feng_d2nt_2023} to generate normals, which are used to enforce depth smoothness supervision.

\subsection{3D GS-based Dense Reconstruction with Pseudo Depth Supervision}\label{sec3.2}
\subsubsection{3D Gaussian Splatting Parameterization}\label{sec3.2.1}
3D-GS has demonstrated robust photo-realistic scene reconstruction with explicit representation and real-time rendering. We follow EndoGS~\cite{endogs} to render RGB images and depth and follow Dai et al.~\cite{surfel} to render normal. Refer to the appendix~\ref{app:3dgs} for detailed parameterization.

\subsubsection{Training with Surface Alignment Constraints}\label{sec3.2.2}
We use the sparse reconstructed priors from OneSLAM (\rev{Sec.~\ref{sec3.1}}) and the keyframe images~\images{K} as the training set and initialize the 3D-GS model with the point cloud~\pointMap~. With the original optimization, Gaussians' parameters are iteratively adjusted to match rendered images to the corresponding ground truth images by minimizing the following photometric loss as described in~\cite{3dgs}:
\begin{align*}
    \loss{pho} &= (1-\lambda_{ssim})\loss{1} + \lambda_{ssim}\loss{D-SSIM}\\
    where~\loss{1} &= \left \| \hat{I} - I \right \|_{1}, ~\loss{D-SSIM} = 1 - SSIM(\hat{I}, I)
\end{align*}
$\hat{I}$ represents rendered image, and $I$ represents ground truth image. 

While the optimization yields realistic scenes when viewed from visited angles, it often becomes trapped in local minima, resulting in noticeable floating artifacts and blending of foreground and background Gaussians when viewed from novel perspectives. This phenomenon is explained by the interdependent nature of Gaussian properties, which means that different configurations of Gaussians can yield identical visual representations, making optimization challenging~\cite {GRM}. This is particularly evident in scenarios like arthroscopy where camera movements are constrained, resulting in inadequate coverage of the target area. To address this, we follow EndoGS~\cite{endogs} for additional geometric constraints $~\loss{d}$ to guide the optimization process to a surface-aligned optimal:
\begin{align*}
    \loss{d} = \left \| \hat{D} - D^{pseudo}\right \|_{1}
\end{align*}
where, $\hat{D}$ represents the rendered depth, and $D^{pseudo}$ represents generated pseudo depth. 

In addition to depth information, normals offer additional constraints that aid in refining the surface geometry. Follow Dai et al.~\cite{surfel} we enforce depth-normal consistency with:
\begin{align*}
    \loss{c} = 1 - \hat{N}\cdot N(\hat{D})
\end{align*}
where $\hat{N}$ is the rendered normal and $N(\cdot)$ converts the depth map to a normal map. Furthermore, we apply the normal-prior regularization to enforce a reasonable surface curvature even under overexposure~\cite{surfel}:
\begin{align*}
    \loss{n} = \lambda_1 (1 - \hat{N}\cdot N^{pseudo}) + \lambda_2 L_1(\triangledown \hat{N}, 0)
\end{align*}
where $\lambda_{i}$ are hyperparameters, $N^{pseudo}$ represents pseudo normal generated from $D^{pseudo}$, and $\triangledown \hat{N}$ represents gradient of rendered normal.

To mitigate the issue of numerous translucent Gaussians overlapping around the surface, we follow~\cite{go} to additionally regularize the opacity of Gaussians, denoted as $\loss{o}$:
\begin{align*}
   \loss{o} = \exp{(\frac{-(o_{i} - 0.5)^2}{0.05})} 
\end{align*}
where $o_{i}$ denotes opacity. This regularization forces Gaussian opacities to become near binary, thus enhancing clarity and reducing visual floating artifacts~\cite{surfel}.

The overall loss is:
\begin{align*}
    \loss{} &= (1-\lambda_{ssim})\loss{1} + \lambda_{ssim}\loss{D-SSIM} + \lambda_{d}\loss{d} + \lambda_{o}\loss{o}\\ 
    &+ \lambda_{c}\loss{c} + \lambda_{n}\loss{n}
\end{align*}
where the hyperparameters $\lambda$ dictate the extent of regularization applied in the optimization process.

\subsection{AR Application Design}\label{sec3.3}
\subsubsection{Measurement Tool}\label{sec3.3.1}
Traditional methods for articular notch measurement during arthroscopy involve inserting a ruler through separate portals, which rely on the surgeon's expertise. Alternatively, preoperative imaging is required to assess the dimensions of the notch and guide the surgical procedure. With the AR measurement tool, surgeons can quickly and efficiently obtain measurements without additional instruments or potential radiation exposure. 
We superimpose the dense 3D-GS model onto the arthroscopic scene. By selecting a point on the rendered image, we unproject the point to 3D and identify its nearest Gaussian neighbors to average the location of the chosen point.
This enables measurement of Euclidean distance between any two points.  

\subsubsection{Annotation Tool}\label{sec3.3.2}
The AR annotation tool 
is implemented in a user-in-the-loop manner. 
We prompt the Segment Anything Model (SAM)~\cite{sam} to generate an initial mask for the region of interest on the arthroscope image. The masked region is then unprojected and intersected with 3D Gaussians. the rendering color of these intersected 3D Gaussians is then modified to highlight critical 3D structures. Alternatively, we can directly anchor 3D shapes onto the 3D model as landmarks.

\section{Experimental Setup}
\subsection{Datasets}
We evaluate our proposed pipeline on two datasets. The First dataset is sampled from a public arthroscopy dataset provided by Marmol et al.~\cite{marmol_dense-arthroslam_2019}. \rev{Video segments with a clear view and pivoting camera motions were extracted from cadaver sequence $H$ with 200 image frames}. The second dataset was generated from an arthroscopy phantom as part of this work and includes \rev{4} video sequences \rev{labeled A, B, C, and D, containing 200, 300, 100, and 200 image frames, respectively (Table~\ref{tab:phantom})}. During the data generation process, a suitable focal length was manually selected and maintained constant throughout each procedure. The motion of the arthroscope is restricted by the insertion portal, permitting rolling, pivoting, forward and backward movement. We also mimicked common clinical motions of the arthroscope to capture the scene.

\begin{table*}[!h]
\caption{Phantom data trials}
\label{tab:phantom}
\centering
\begin{adjustbox}{width=1.0\textwidth}
\begin{tabular}{c|c|c|c|c|c|c|c}
\toprule
\textbf{Trials} &\textbf{Target} &  \textbf{Motion} &\textbf{Entry} &\textbf{Anatomy} &\textbf{Duration (s)} &\textbf{Num Frames} &\textbf{Used Frames Range}\\
\midrule
A & 1517-29-2 & Common motion & Left entry & meniscus, articular cartilage, femur, patella, cruciate ligament & 49.1 & 1228 & 1-200\\
B & 1517-29-2 & Pivoting & Left entry & meniscus, articular cartilage, cruciate ligament & 37.6 & 939 &1-200\\
C & 1517-29-2 & Pivoting & Left entry &  femur, cruciate ligament & 36.4 & 909 &1-300\\
D & 1517-29-2 & Forward/Backward & Right entry & meniscus, articular cartilage, cruciate ligament, femur, patella & 37.2 & 929 &100-300\\
H & Knee Cadaver & Common motion & - & meniscus, articular cartilage, cruciate ligament & 52.6 & 1578 &1-200\\
\bottomrule
\end{tabular}
\end{adjustbox}
\end{table*}

\subsection{Physical Setup}

\begin{figure}[!t]
\centering{\includegraphics[width=8cm]{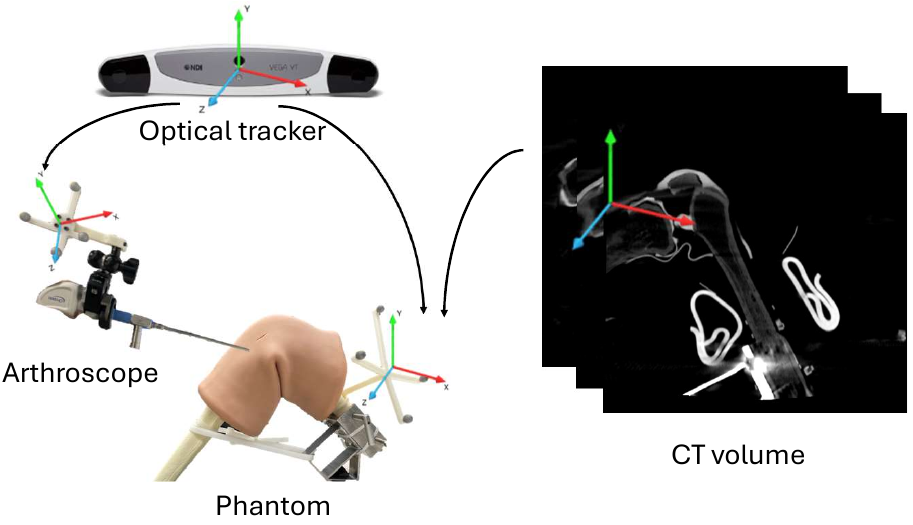}}
\caption{The physical setup for phantom data collection. Fiducial markers are rigidly attached to the arthroscope and arthroscopy phantom. The 6 DoF poses are tracked by an optical tracker during inspection. The CT scan is taken after data collection to gain a ground truth model.}
\label{fig:expSetup}
\end{figure}

The phantom data collection setup is visualized in Fig.~\ref{fig:expSetup}. We use an optical tracker to capture the ground truth poses of the arthroscope (Stryker 502-477-031\footnote{https://www.stryker.com/us/en/portfolios/medical-surgical-equipment/surgical-visualization/scopes.html}) and knee phantom (SAWBONES KNEE ARTHROSCOPY 1517-29-2\footnote{https://www.sawbones.com/knee-arthroscopy-w-normal-meniscus-patella-patella-tendon-bone-clamp-c-clamp-movable-from-0-extension-to-120-flexion-1517-29-2.html}). Optical markers are rigidly attached to the shaft of the arthroscope and the phantom. We perform a camera calibration by moving the arthroscope while looking at a fixed ChArUco Board with optical markers poses tracked. Through Zhang's method~\cite{zhang2000flexible}, we find the camera intrinsic parameters. We further perform a hand-eye calibration routine~\cite{horaud1995hand, furrer2018evaluation} to obtain hand-eye transformation between the arthroscope center and optical markers. To obtain the ground truth 3D model, we take preoperative CT scans with LoopX (BrainLab~\footnote{https://www.brainlab.com/surgery-products/overview-platform-products/robotic-intraoperative-mobile-cbct/}). 


\subsection{3D Reconstruction Evaluation}
We first evaluate the accuracy of our reconstruction. Since the sparse reconstruction relies solely on monocular images, and the dense model is built on top of the sparse 3D priors, it inherently encounters scale ambiguity. Therefore, we align the reconstructed point cloud from the 3D GS model at scale with the ground truth model using the Iterative Closest Point (ICP) algorithm~\cite{icp}. Since the 3D-GS model is explicitly represented as a dense point cloud, it is evaluated using a point-to-point distance approach. 
The ground truth mesh is first converted into a point cloud by sampling points from its surface. Then nearest neighbors on the ground truth are found for each point in the reconstructed point cloud as correspondences. We then calculate the root mean squared error (RMSE) between correspondences to quantify the overall reconstruction error, and the Hausdorff Distance, providing insights into overall shape similarity and specific areas of deviation. Additionally, we assess the quality and similarity of rendered images to ensure high fidelity. High-fidelity images preserve the visual integrity and details of the original scenes, which is particularly beneficial for clinical AR applications. We especially choose Peak Signal-to-Noise Ratio (PSNR) to quantify the pixel-level errors and Structural Similarity Index Measure (SSIM) for image similarity in human visual perceptual aspect.
Using these evaluation metrics, we compare our proposed reconstruction pipeline with the classical Structure-from-Motion (SfM) method COLMAP~\cite{schoenberger2016sfm, schoenberger2016mvs}, a differentiable camera pose and 3D geometry estimation method FlowMap~\cite{flowmap}, and a ViT-based 3D reconstruction paradigm DUSt3R~\cite{dust3r}. For COLMAP~\cite{schoenberger2016sfm, schoenberger2016mvs}, we use a simple radial model with default initial parameters. For FlowMap~\cite{flowmap} and DUSt3R~\cite{dust3r}, we use the pre-trained model and keep default parameters. 

\subsection{Articular Structure Measurement Accuracy}
The measurement process is simulated by randomly selecting $500$ pairs of points on the reconstructed scene and calculating the Euclidean distance between them. The corresponding points on the ground truth model are then identified to calculate the ground truth distance. The effectiveness of our measurement tool is evaluated by analyzing the distribution of the distance errors and comparing it against the state-of-the-art models.

\subsection{Annotation Anchoring Evaluation}
The accuracy of annotation anchoring is assessed by benchmarking our AR application against Cutie~\cite{cutie}: a state-of-the-art video object segmentation method, on the public cadaver sequence $H$. Employing Cutie, we deploy point prompts to assist in segmenting the cruciate ligament from the initial endoscope video frame. The resulting mask is then propagated to subsequent frames to estimate plausible masks. In contrast, in our approach, we use the same initial mask to delineate the designated region and project it onto all image frames during real-time rendering. We then compare the mean Intersection over Union (mIoU) between these two sets of segmentation masks.

\begin{table*}[!t]%
\caption{3D Reconstruction Results. We evaluate our reconstruction accuracy and fidelity relative to COLMAP~\cite{schoenberger2016sfm, schoenberger2016mvs}, Flowmap~\cite{flowmap} and Dust3R~\cite{dust3r} on four datasets. Our method outperformed or is comparable to these methods.}
\centering
\label{tab:reconstruction}
\scalebox{0.7}{
\begin{tabular*}{\textwidth}{@{\extracolsep\fill}llllll@{\extracolsep\fill}}%
\toprule
\multirow{2}{*}{\textbf{Trials}} & \multirow{2}{*}{\textbf{Methods}} & \textbf{RMSE} &\textbf{Hausdorff} & \multirow{2}{*}{\textbf{PSNR}} & \multirow{2}{*}{\textbf{SSIM}} \\
& & \textbf{(mm)} & \textbf{Distance(mm)} & & \\
\midrule
\multirow{4}{*}{A}&  COLMAP &       3.00 &   59.4 &     - &    0.15 \\
      & Flowmap &    \textbf{1.22} &   8.02 &    29.34 &    0.66 \\
      &  Dust3R &    1.59 &   \textbf{7.11} &    28.31 &    0.72 \\
      &    \textbf{Ours} &    3.79 &  15.23 &    \textbf{30.21} &    \textbf{0.82} \\
\midrule
\multirow{4}{*}{B} &  COLMAP &    3.07 &  37.69 &     - &    0.16 \\
      & Flowmap &   11.48 &  54.24 &    28.79 &     0.70 \\
      &  Dust3R &    1.78 &   \textbf{9.48} &    29.11 &    0.72 \\
      &    \textbf{Ours} &    \textbf{1.52} &   9.58 &    \textbf{32.98} &    \textbf{0.90} \\
\midrule
\multirow{4}{*}{C} &  COLMAP &    1.94 &  12.75 &     - & 0.27 \\
 & Flowmap & 1.95 &9.69 & 29.04 & 0.64 \\
&  Dust3R & 1.83 &7.19 & 28.86 & 0.74 \\
& \textbf{Ours} & \textbf{1.57} &\textbf{7.08} & \textbf{33.77} &  \textbf{0.90} \\
\midrule
\multirow{4}{*}{D} &  COLMAP & \textbf{1.31} &  44.35 &  - & 0.25 \\
& Flowmap & 2.35 &17.2 & 29.35 & 0.66 \\
&  Dust3R &  1.50 &\textbf{5.89} & 29.54 & 0.77 \\
& \textbf{Ours} & 1.96 &7.06 & \textbf{34.51} & \textbf{0.94} \\
\midrule
\multirow{4}{*}{Average}&  COLMAP & 2.33 &  38.54 &-&  0.20 \\
& Flowmap & 4.25 &22.2875 & 29.13 &0.66 \\
&  Dust3R &\textbf{1.675} & \textbf{7.4175} &28.95 &  0.73 \\
& \textbf{Ours} & 2.21 & 9.7375 &  \textbf{32.86} & \textbf{0.89} \\
\bottomrule
\end{tabular*}}
\end{table*}

\section{Results and Discussion}
\subsection{Results in 3D Reconstruction}
\label{sec:recon}
Our proposed method exhibits comparable or improved performance relative to COLMAP~\cite{schoenberger2016sfm, schoenberger2016mvs}, Flowmap~\cite{flowmap}, and Dust3R~\cite{dust3r} in both RMSE and Hausdorff Distance (Table~\ref{tab:reconstruction}), indicating a robust capability to reconstruct the 3D geometry of the scene. Furthermore, our method surpasses all other techniques in PSNR and SSIM metrics, highlighting its superior capability for high-fidelity rendering. Refer \rev{to} Table~\ref{tab:p} in \rev{the} appendix for statistical P-values. 

COLMAP~\cite{schoenberger2016sfm, schoenberger2016mvs} and Flowmap~\cite{flowmap} perform adequately in certain trials, such as \rev{$D$}, where the cameras move back and forth in feature-rich areas. However, the reconstruction quality is highly inconsistent, particularly when there are significant changes in illumination or when the arthroscope moves towards feature-scarce regions like the femur bone. This instability in feature tracking leads to errors in mapping specific regions, causing an increased Hausdorff distance. We avoid calculating the PSNR for COLMAP~\cite{schoenberger2016sfm, schoenberger2016mvs} because, even after MVS dense reconstruction, the camera view scenes remain sparse. The PSNR calculations can be skewed by large empty regions, inaccurately reflecting the quality of the reconstruction. The low SSIM value for COLMAP~\cite{schoenberger2016sfm, schoenberger2016mvs} indicates a noticeable difference between the real scene and the reconstructed scene, which humans are likely to perceive as dissimilar. While Flowmap~\cite{flowmap} can generate dense scenes with good rendering fidelity, it could still fail to produce accurate surfaces. For instance, in trial $C$, point maps fail to merge correctly due to inconsistent depth estimation, resulting in outliers that significantly increase geometric errors. 

Dust3R generates high-quality 3D geometry with fewer outliers, due to its pixel-by-pixel point map alignment. As Dust3R~\cite{dust3r} reconstructs scenes in an end-to-end manner and relies on the point map to estimate camera parameters and poses, it can hinder the rendered image from having an identical viewing angle to the real image.

\begin{figure}[!t]
\centering{\includegraphics[width=0.48\textwidth]{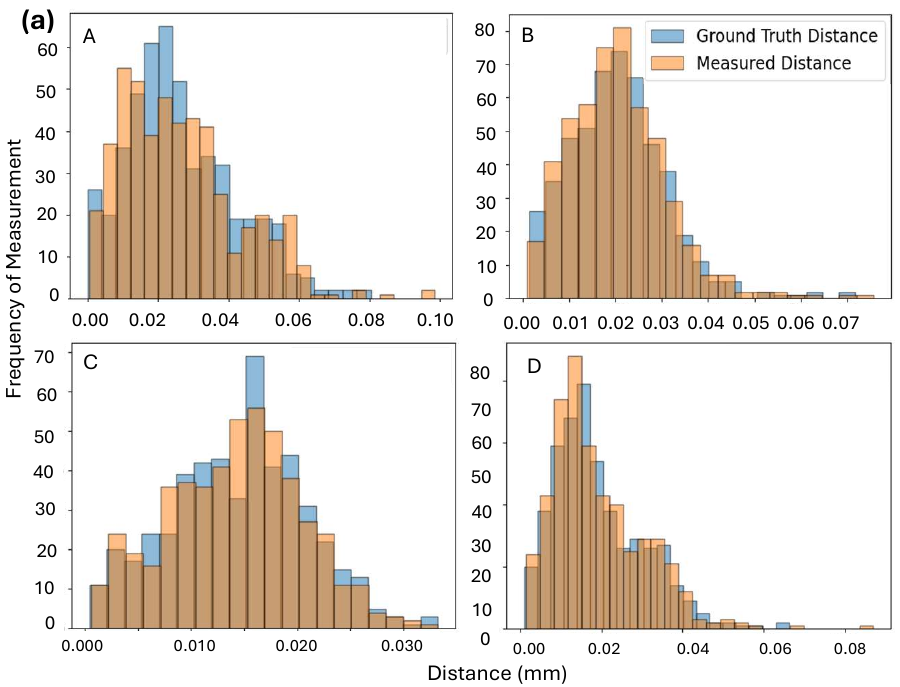}}
\centering{\includegraphics[width=0.48\textwidth]{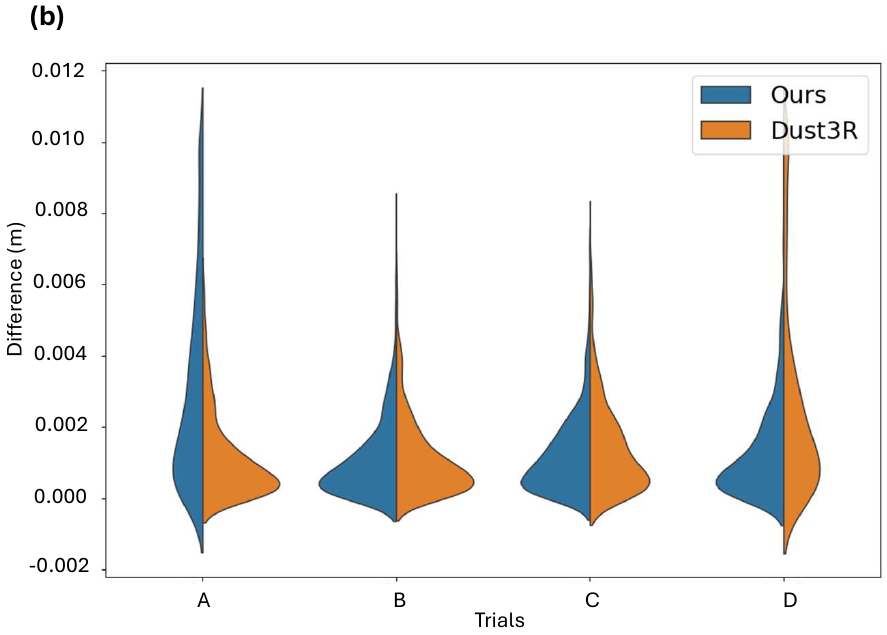}}
\centering{\includegraphics[width=0.48\textwidth]{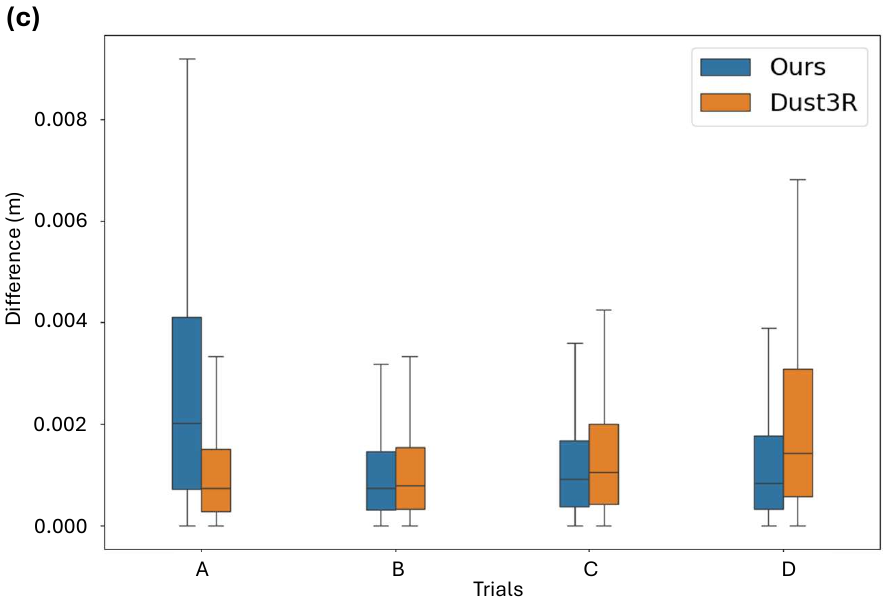}}
\caption{Evaluation results for AR Measurement accuracy. (a) Distribution alignment between the measured distance on our reconstruction (orange), and the ground truth distance (blue) for the evaluation trials. (b) \rev{Comparison of the distribution shapes reflecting measurement difference between our method (blue) and Dust3R (orange)}. \rev{(c) Comparison of the key statistical indicators.}}
\label{fig:meas_eval}
\end{figure}

Our approach jointly optimizes photometric similarity and the 3D geometry of the scene. The relative scale recovery effectively ensures consistency in monocular depth estimation across frames, providing geometric information that guides the convergence of the 3D GS model and \rev{make} it better aligns with the real target surface. Additionally, the use of normals further improves surface smoothness. Despite the efficiency of our method, it still encounters issues. In trials $A$ and \rev{$D$}, cavity areas with less illumination lead to false depth estimation, causing significant deviations in those regions and adversely affecting the overall alignment. These results are reasonable, considering the monocular depth estimation model has never been trained on arthroscopy video before. In the future, we will fine-tune the depth model to overcome this issue.

\begin{figure}[!t]
\centering{\includegraphics[width=0.48\textwidth]{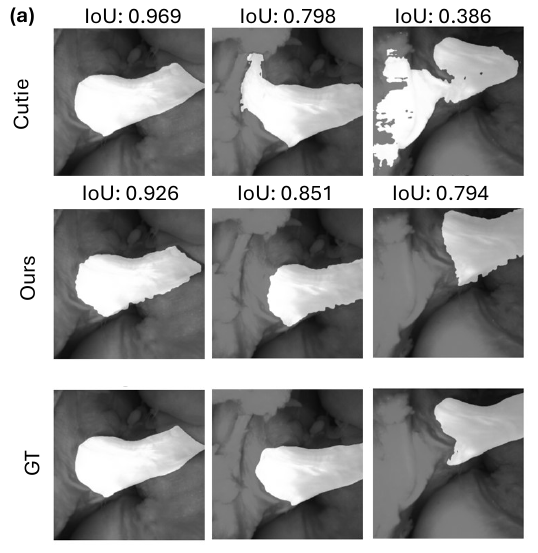}}
\centering{\includegraphics[width=0.48\textwidth]{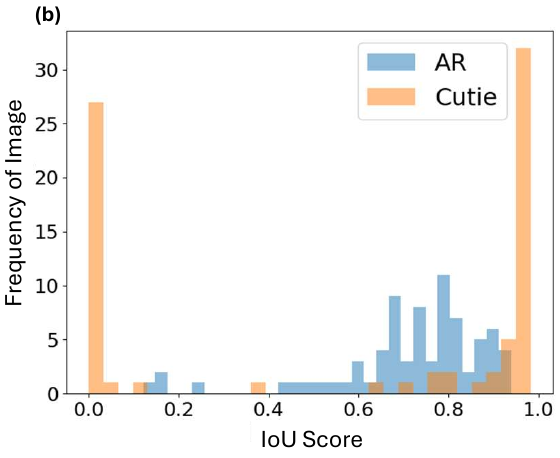}}
\caption{Evaluation results for AR annotation accuracy. (a) Qualitative results for AR annotation accuracy. The annotation masks are highlighted and superimposed on the arthroscopic scene with the corresponding IoU score on top of each method. Our pipeline is comparable with the SOTA segmentation method on initial frames and outperforms it for consecutive frames. (b) The histogram of IoU score for AR annotation and Cutie segmentation. The results of Cutie (orange) are polarized, showing good segmentation on some frames but losing tracking on others. In contrast, our method (blue) achieves relatively steady and consistent annotation.}
\label{fig:ar_seg}
\end{figure}

\subsection{Results in Measurement Accuracy}
Overall, our measurement application achieves result of $1.59 \pm 1.81\text{mm}$ error on average in articular notch measurement and shows potential feasibility for clinical use. As shown in Figure~\ref{fig:meas_eval}(a), the distribution of measurement results on our reconstruction and the ground truth model are similar in shape, central tendency, and overall pattern. This implies that our reconstruction accurately represents most of the true geometry of the scene. 
\rev{Regarding the accuracy in distances, our method achieves comparable results to Dust3R in trial $A$ and performs better in $B$, $C$, and \rev{$D$}. The violin plot Fig.~\ref{fig:meas_eval}(b) and box plot Fig.~\ref{fig:meas_eval}(c) illustrate the distribution of measurement errors for our method and Dust3R across different trials. The violin plot shows the distribution shape, while the box plot provides key statistical insights such as median and quartiles. In Trial $A$, our method shows a longer tail, indicating a higher variability in measurements. This is attributed to a slightly less accurate reconstruction with outliers, as detailed in Sec.~\ref{sec:recon}. For Trial $B$, $C$, and $D$, our method outperforms Dust3R, with a more concentrated distribution of measurement errors around the median, reflecting better accuracy and reliability.}

\subsection{Results in Annotation Accuracy}
Our AR annotation application achieves $\text{mIoU} = 0.721$, competitive to Cutie with $\text{mIoU} = 0.569$, and our application is significantly better in accurately annotating the ligament than Cutie proved by a p-value $<0.001$ over IoU scores. As shown in~\ref{fig:ar_seg} (a), we use the ground truth mask at the first column as initial information for both methods. Even though Cutie has decent performance on images with similar camera poses~\ref{fig:ar_seg}(b), it fails on consistently predict annotation throughout the sequence as examples shown in the second and third columns. Especially in frames where the ligament regions are small or with illumination variation, the annotation is propagated to random regions or vanishes, causing a confusing annotation. In contrast, our AR application highlights each \rev{Gaussian} as the anchor, successfully maintaining the geometric information in 3D space for a more consistent annotation. The inaccuracy of our annotation is related to the error of reconstruction, moreover, the superimpose quality is sensitive to camera intrinsic parameters.

\section{Limitations}
Our pipeline is limited by the quality of point tracking. Since our SLAM algorithm relies on image correspondences, it is sensitive to inaccurate point estimation. For instance, points sampled on the joint surface can drift over time, introducing noise into the point map and thereby compromising the scale recovery process. 

Moreover, the monocular depth estimation exhibits inconsistent predictions for relative depth in anatomical structures across different frames, especially for cavity regions. Additionally, our pipeline currently assumes a rigid scene and fixed camera intrinsic parameters throughout the procedure. Therefore, the feasibility of the proposed pipeline in a real clinical environment requires further refinement.

\section{Conclusion}\label{sec6}
In this work, we propose and evaluate a vision-based arthroscopy scene reconstruction technique and explore the integration of augmented reality (AR) applications. In our proposed method, we leverage \rev{the} vision-based SLAM algorithm to obtain sparse 3D priors and combine it with pseudo monocular depth information to reconstruct the articular scene using a 3D GS model. Utilizing this 3D model, we developed an AR surgical guidance application featuring AR annotation and measurement tools, aiming to simplify procedures and provide more convenient assistance for surgeons. We demonstrate that our pipeline generates superior reconstructions compared to three widely used methods and offers more robust measurement and annotation capabilities than the recent state-of-the-art approach. In the future, we plan to further improve our pipeline by addressing its current limitations. We aim to incorporate a low-cost external tracking strategy to complement vision-based localization, enhancing the overall reliability and clinical feasibility.



\bibliographystyle{ieeetr}
\bibliography{reference}

\clearpage
\section{Supplementary}
\subsection{3D GS Preliminaries}\label{app:3dgs}
Each 3D Gaussian function $\{G_{i}\}_{i=1}$, representing a point, is characterized by a covariance matrix $\Sigma$ and centered at the mean $\mu \in \mathbb{R}^{3}$. $\Sigma$ is parameterized as scaling matrix $S \in \mathbb{R}^{3}$ and rotation matrix $R$ represented by unit quaternion $q \in \mathbb{R}^{4}$. Point-based $\alpha$-blending is used to render color for pixel $p$:
\begin{align*}
    C(p) &= \sum_{i \in N}c_{i} \alpha_{i}\prod_{j=1}^{i-1} (1-\alpha_{j})\\
    where ~c_{i} &= SH(sh_{i}, v_{i}),~\alpha_{i} = \sigma_{i} G_{i}
\end{align*}
The Sphere Harmonics (SH) is used to formulate color for view direction $v_{i}$ with SH coefficient $sh_{i}$, and $\sigma_{i}$ means the opacity. \rev{The} pixel-wise depth and normal can be rendered in a similar manner:
\begin{align*}
    D(p) &= \sum_{i \in N}z_{i} \alpha_{i}\prod_{j=1}^{i-1} (1-\alpha_{j}), ~N(p) = \sum_{i \in N}R^z_{i} \alpha_{i}\prod_{j=1}^{i-1} (1-\alpha_{j})
\end{align*}
Where $z_{i}$ is the distance from the camera center to the Gaussian center $\mu_{i}$, $R^z_{i}$ is the unit vector along the z-axis after the rotation.
Overall, each Gaussian can be parameterized as $G_{i}(\mu_{i}, q_{i}, s_{i}, sh_{i}, \sigma_{i})$.

\subsection{Experimental Platforms}
The reconstruction method proposed is executed on a local Linux server equipped with an AMD Ryzen 7 3800X CPU, RTX TITAN GPU, running Ubuntu 20.04 LTS. The development of AR rendering and applications takes place on a laptop featuring an Intel i7-11800H CPU, and RTX 3060 Laptop GPU, and is deployed using Unity 2022.3.7f1 on a Windows 11 operating system.

\begin{table*}[!b]%
\caption{P-value for 3D Reconstruction Results compared to our method. For PSNR and SSIM our method is significantly better than all other methods, while performs comparable on RMSE and Hausdorff distance.}
\centering
\label{tab:p}
\scalebox{0.7}{
\begin{tabular*}{\textwidth}{@{\extracolsep\fill}lllll@{\extracolsep\fill}}%
\toprule
 \multirow{2}{*}{\textbf{Methods}} & \multirow{2}{*}{\textbf{RMSE}} &\textbf{Hausdorff} & \multirow{2}{*}{\textbf{PSNR}} & \multirow{2}{*}{\textbf{SSIM}} \\
& & \textbf{Distance} & & \\
\midrule
  COLMAP &       \textcolor{red}{0.43\%} &   \textcolor{blue}{0.01\%} &     - &    \textcolor{blue}{< 0.01\%} \\
Flowmap &    \textcolor{red}{0.22\%} &   \textcolor{red}{0.14\%} &    \textcolor{blue}{< 0.01\%} &    \textcolor{blue}{< 0.01\%} \\
 Dust3R &    \textcolor{red}{0.18\%} &   \textcolor{red}{0.15\%} &    \textcolor{blue}{< 0.01\%} &    \textcolor{blue}{< 0.01\%} \\
\bottomrule
\end{tabular*}}
\end{table*}

\subsection{\rev{Computation Evaluation}}
\rev{Table \ref{tab:ct} shows the average reconstruction time of each method for all four arthroscopy sequences. We observe that the proposed pipeline has significantly superior performance compared to COLMAP and Flowmap. Limited by our experimental platform, we employ only 10 images for Dust3R to reconstruct each scene, compared to 200 images for other methods, resulting in faster execution times but compromised quality.}
\begin{table*}[!b]%
\rev{
\caption{Computational time comparison.}
}
\centering
\rev{
\label{tab:ct}
\scalebox{0.7}{
\begin{tabular*}{\textwidth}{@{\extracolsep\fill}ll@{\extracolsep\fill}}%
\toprule
 \textbf{Methods} & \textbf{Total Time in Average (s)}\\
\midrule
  Ours &   424.25\\
  COLMAP &  872.865\\
  Flowmap &  1817.98\\
 Dust3R &   63.7\\
\bottomrule
\end{tabular*}}
}
\end{table*}

\begin{figure*}[!b]
\centerline{\includegraphics[width=\textwidth]{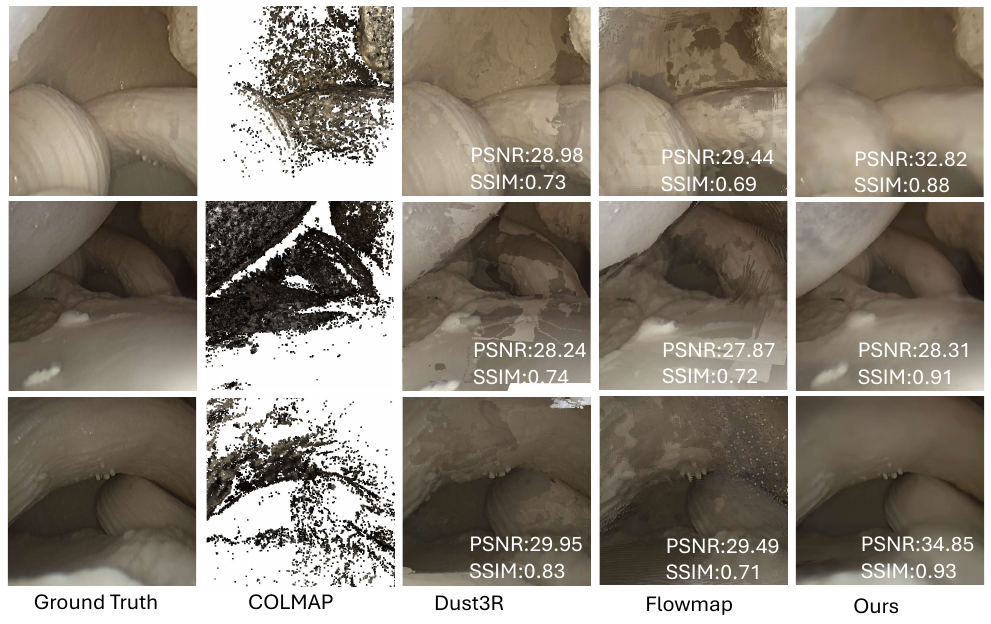}}
\caption{Visualization of rendering quality comparison.}
\source{
\begin{flushleft}
\end{flushleft}
}
\end{figure*}
\end{document}